\documentclass[a4paper]{article}
\usepackage[dvipsnames]{xcolor}

\usepackage{INTERSPEECH2021}
\usepackage{tipa}
\usepackage{algpseudocode}
\usepackage{algorithm}
\usepackage{lipsum}
\usepackage{subfig}
\algnewcommand\algorithmicforeach{\textbf{for each}}
\algdef{S}[FOR]{ForEach}[1]{\algorithmicforeach\ #1\ \algorithmicdo}

\title{Spoken Term Detection Methods for Sparse Transcription\\ in Very Low-resource Settings}
\name{\'Eric Le Ferrand$^{12}$, Steven Bird$^1$, Laurent Besacier$^2$}
\address{
  $^1$Charles Darwin University,
  $^2$Universit\'e Grenoble Alpes}
\email{name.surname@cdu.edu.au, name.surname@univ-grenoble-alpes.fr}

\begin{document}

\maketitle
\begin{abstract}
% This paper explores a new method known as sparse transcription for transcribing speech collections in under-resourced contexts using word spotting methods.
% We compare the performance of two such methods, dynamic time warping and phone recognition, in two very low resource languages: Kunwinjku and Mboshi.
% We show that a few minutes of transcribed speech can be used to fine-tune an acoustic model to obtain an approximate transcription.
% The resulting noisy phone sequence can then be processed using string matching to
% sparsely transcribe our speech collections, yielding results which outperform dynamic time warping in terms of precision.
% Finally a parsing across a confusion network of phonemes output by the acoustic model using a written lexicon stored in a trie allows us to boost the recall, while maintaining high precision.
We investigate the efficiency of two very different spoken term detection approaches for transcription when the available data is insufficient to train a robust ASR system.
This work is grounded in very low-resource language documentation scenario where only few minutes of recording have been transcribed for a given language so far.
Experiments on two oral languages show that a pretrained universal phone recognizer, fine-tuned with only a few minutes of target language speech, can be used for spoken term detection with a better overall performance than a dynamic time warping approach.
In addition, we show that representing phoneme recognition ambiguity in a graph structure can further boost the recall while maintaining high precision in the low resource spoken term detection task.

\end{abstract}


\section{Introduction}
Efforts are made across Australia to preserve, document and revitalize Aboriginal languages. These languages exist primarily in spoken form, and even if there is an official orthography it is not widely used by local people.
Making recordings of speakers has been a widespread practice for documenting traditional knowledge.
However, such recordings are often not transcribed, making them hard to access.

Manual transcription is time consuming and is often described as a bottleneck \cite{brinckmann2009transcription}.
While automatic speech recognition (ASR) has seen great improvement in recent years \cite{Povey_ASRU2011,watanabe2018espnet}, it relies on a large amount of annotated data. Attempts to build ASR systems for low-resource languages end up with high word error rate or single-speaker models making them of limited use in indigenous contexts \cite{gupta2020automatic, gupta2020speech,foley2018building, adams2020user}.
%\cite{foley2018building, adams2020user} propose computer assisted way to support speech transcription, but their methods still rely on ASR.

Such methods assume that everything should be transcribed.
\cite{bird2020sparse} describes a sparse transcription model where we only transcribe the words we can confidently recognise, using word-spotting, while leaving the transcription of more difficult sections for later, perhaps when a speaker is available \cite{bird2020sparse}.
Based on this model, we proposed a workflow which combines spoken term detection and human-in-the-loop to support transcription in under-resourced settings \cite{le2020enabling}.
Here, we focus on the spoken term detection stage of the workflow, trying to seek a higher-precision method that better supports interactive transcription.

Automatic phone recognition has seen progress with minimal data \cite{gupta2020speech, li2020universal}. While \cite{bird2020sparse} argues that phonetic transcriptions do not stand in for the speech data and cannot be segmented to generate the required higher-level word units,
we can nevertheless use phone transcriptions as a speech encoding, retaining our commitment to the sparse transcription model.
In this paper we show how this can be done, and compare it with another spoken term detection method, namely dynamic time warping (DTW) \cite{sakoe1978dynamic}.
We consider both methods as applied to two very low-resource languages, Kunwinjku (gup) and Mboshi (mdw).

\section{Background}
\label{sec:context}

Several sources of annotated data can be used to support transcription.
Combining automatic processes and human intervention can be a way to iteratively fine-tune a speech recognition model with newly produced data, and enhance the manual transcription with automatically annotated data. \cite{bird2020sparse} proposes an approach to transcription which combines word spotting with human-in-the-loop in a iterative process.
%This new model consists into a set of tasks including manual lexicon building, spoken term detection and manual contiguous transcription. 
\cite{le2020enabling} explore this further, using DTW with basic speech representations to transcribe up to 42\% of a lexicon in their speech collection.
However, this method is not robust in the face of speaker variability.
Research around speech features for spoken term detection has explored the use of bottleneck features, or the hidden representation of an AutoEncoder \cite{menon2019feature, kamper2015unsupervised}, although they make negligible difference in the context of the workflow in \cite{le2020enabling}.
Others have exploited neural approaches to train word classifiers from words pair using Siamese loss \cite{settle2016discriminative,settle2017query}, however pairs of words are required, limiting the selection of words that can be searched.
    
While standard automatic speech recognition methods tend to give poor performance in low-resource setting \cite{gupta2020automatic, gupta2020speech},
recent work has shown that it is possible to obtain an automatic phone transcription with minimal training \cite{gupta2020speech, li2020universal, adams2018evaluating}. 
In the context of word-level transcription, \cite{bird2020sparse} describes phone transcription as a `retrograde step'
given that optimising the recognition of low-level units such as phones does not necessarily help with identifying higher-level units.
%%%%%%%%%%%%%%%%%%%%ERIC CHANGED SOMETHING%%%%%%%%%%%%%%%%%%%%%%%%%%
He also observes that only linguists can provide the required phone transcriptions or graph-to-phone rules.
In many spoken languages, we can find standard orthographies and a small amount of spoken material transcribed. These orthographies are built in a written system close to the language phonology. Such system is documented enough so an approximate phone transcription can be easily obtained through a basic mapping from graphemes to phonemes on existing written data.
%%%%%%%%%%%%%%%%%%ERIC CHANGED SOMETHING%%%%%%%%%%%%%%%%%%%%%%%%%%%%%
\cite{michaud2018integrating} explain that the linguist should aim for a faithful phones transcription to aim to a better Phone Error Rate (PER). However a noisy phones transcription or even an orthographic transcription mapped into phonemes can also be used to train a phone recognizer. Phone transcription has this advantage over speech representation that its level of abstraction gives a speaker-independent representation. 
    
Doing word spotting, while some terms are unlikely to be found in a noisy phone transcription, doing exact matching between the written forms of a lexicon (transliterated into phones) and a stream of the k-best phones output by the recogniser offers inter-speaker robustness.
%makes sense in a very-low resource context when a few minutes of transcribed speech distributed across a few speakers is the only data available.

In the context of textual data annotation, \cite{felt2012improving} argues that an automatic annotation is likely to improve the speed of the annotator if the precision is at least 80\% and the accuracy if it is at least 60\%. Even though a wide set of factors can have an influence on the speed and accuracy, like the quality of the data or the confidence of the transcriber, we argue that a higher precision is to be prioritized over a higher recall. 
% In the early stage of transcription with members of an indigenous community, one of the first challenge is to make them familiar with new kinds of task. \cite{le2020enabling} introduce an interactive activity in which the speakers are supposed to manually verify the output of a spoken term detection system. In this context, a high precision is to be privileged over a high recall to more easily engage with the community.\todo{TODO: say that it was not successful so we look for a better way to have high precision ?}

\section{Methods}

We begin with a lexicon of size \textit{s} consisting of audio clips of spoken words, along with their orthographic transcription,
and a speech collection in which more instances of those words may be found. 
%We use a word-spotting method to retrieve speech terms in the collection that match those in the lexicon. \textit{n} possible matches are the presented to a speaker for verification. We simulate this workflow with two
Two spoken term detection approaches are investigated here:
(a)~a baseline method based on DTW applied on MFCC features (normalised for cepstral mean and variance); and
(b)~a method based on automatic phone recognition in which phones units are mapped to word units (P2W).

\subsection{Baseline: Sparse Transcription using DTW}

The baseline is a simplified version of
 the workflow of \cite{le2020enabling} which introduced an iterative process to transcribe speech  with a growing spoken lexicon. 
We use DTW to retrieve lexical items from the speech collection, sliding query terms across the utterances.
%This process is applied through a single iteration (as opposed to several iterations in \cite{le2020enabling}).

% For each entry in the lexicon, based on the DTW score, we select the best 50 candidates for Mboshi and 30 for Kunwinjku to be presented to a speaker for verification. No additional threshold is applied.
For each entry in the lexicon, based on the DTW score, we select the $n$ best matches in the collection to be evaluated.
%per word to be verified. 
The value of $n$ has been optimized using a development set ($n=5$ for Kunwinjku and $n=20$ for Mboshi).

\subsection{Sparse Transcription using Phone Recognition (P2W)}

\begin{figure}[!ht]%
\centering
    \includegraphics[width=0.6\linewidth]{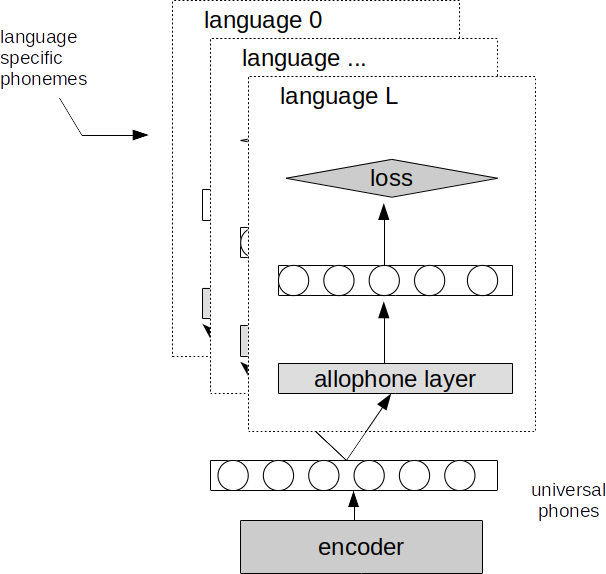}%
    \caption{Allosaurus model \cite{li2020universal}}%
    \label{fig:allo}%
\end{figure}

\cite{li2020universal} introduced Allosaurus, a universal phone recognition system which combines a language independent encoder and phone predictor, and a language dependent allophone layer and a loss function associated with each language (Fig.~\ref{fig:allo}). 
%The encoder first produces the phones distribution, then the allophone layer transforms the phones distribution into a phonemes distribution for each language. The Allosaurus model has been trained over 2000 languages using the PHOIBLE database \cite{moran2014phoible}.
%it incorporates knowledge of languages' phonology into its multilingual model through an allophone layer. This layer maps a universal narrow phone set with the phonemes that appear in the transcription of each language. 
Allosaurus models are trained using standard phonetic transcriptions and the allovera database \cite{mortensen2020allovera}, a multilingual allophone database that can be used to map allophones to phonemes.
The model first encodes speech with a standard ASR encoder which computes the universal phone distribution.
Then an allophone layer is initialized with the allophone matrix and maps the universal phone distribution into the phoneme distribution for the given target language.
The resulting model can be fine-tuned and applied to unseen languages.

% \begin{figure}[!ht]%
%     \centering
%     \subfloat{{\includegraphics[width=0.45\textwidth]{training_allo.png} }}%
%     \hspace{-5mm}
%     \subfloat{{\includegraphics[width=0.5\textwidth]{valid_allo.png} }}%
%     \caption{Ugly illustration of the workflow}%
%     \label{fig:pipeline}%
% \end{figure}
\begin{figure}[!ht]%
    \centering
    \includegraphics[width=0.75\linewidth]{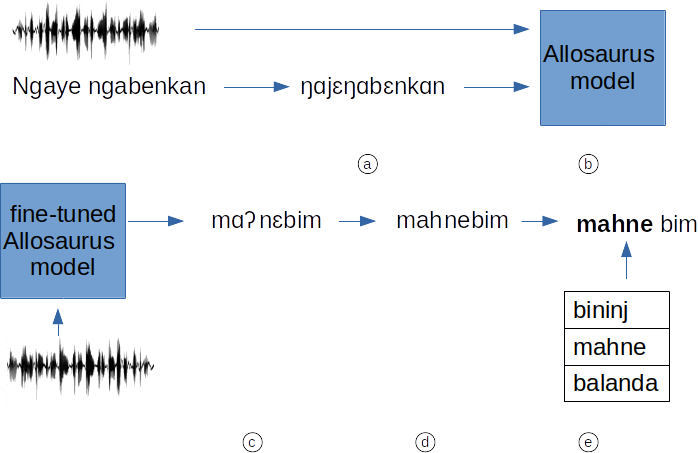}%
    \caption{Illustration of P2W workflow}%
    \label{fig:pipeline}%
\end{figure}

In the current context, since we only have an orthographic transcription for Kunwinjku, we transliterate it into IPA (Fig.~\ref{fig:pipeline}a) with the mapping shown in Table~\ref{mapping}.
The transcription contains some English words which will be mapped as if they were Kunwinjku words (eg., school is written /s\textipa{P}kool/ instead of /sk\textupsilon l/).
For Mboshi, the orthographic transcription already mostly matches the corresponding phonetic transcription.\footnote{The tones are marked in the orthographic transcription but this feature is not taken into account in the Allosaurus model. We thus decided to treat the orthographic transcription as a phonetic transcription so the accentuated vowels are considered as new phones.}

\begin{table}[!h]
\setlength{\tabcolsep}{0.5\tabcolsep}
\begin{tabular}{l|lllllllllllllllll}
graphs &a  & b & d & h  & e  & i  & ch & y  & o   & k & dj & s & r & rr  \\
phones & \textscripta  & b & d & \textglotstop  & \textepsilon   & i  & \textesh  & j  & \textopeno   & k  & \textbardotlessj  & s & \textturnrrtail & r   \\ \hline
graphs & ng & rd & rl & nj & rn & u & f & l & m & n & w & p & t     \\
phones & \textipa{ŋ}  & \textrtaild  & \textrtaill  & \textltailn  & \textrtailn  & u & f &l & m & n  & w & p & t \\
\end{tabular}
\caption{Grapheme to phoneme mapping for Kunwinjku}
\label{mapping}
\end{table}

We fine-tuned the original pretrained model with the training subsets described in Section \ref{sec:data} following the mapping described above resulting on one new phone recognition model per language (Fig.~\ref{fig:pipeline}b). We used the resulting models to automatically transcribe the validation set of the two languages (Mboshi and Kunwinjku) (Fig.~\ref{fig:pipeline}c).
For Kunwinjku, we reverse the mapping function to get a grapheme transcription (Fig.~ \ref{fig:pipeline}d).
Finally, we use the written lexicon for each language and perform spoken term detection using the queries' transcripts using a longest string matching algorithm (Fig.~\ref{fig:pipeline}e).
%parse it through the transcription automatically generated to match same forms. 
%If two different words are found in the same area, the longest word is kept.  

%%%%%%%%%%%%%%%%%%%%%%%%%%%%%%%%%%%%%%%%%%%%%%%%%%%%%%%%%%%%%%%%%%%%%%%%%%%%%%%%
\section{Data}
\label{sec:data}
The same datasets of \cite{le2020enabling} will be used: a corpus in Kunwinjku which consists of 301 utterances aligned with an orthographic transcription and a forced alignment created using the MAUS forced aligner \cite{kisler2017multilingual},
and a corpus in Mboshi which consists of 5130 utterances elicited from text, with orthographic transcription and a forced alignment at the word level \cite{godard2017very}.

In order to train the phone recognition system, an additional collection of 20 min of Kunwinjku (newly recorded guided tours) is used as training set.
For Mboshi, a subset of 30 min is randomly extracted from the original corpus.

The lexicon queries (for spoken term detection) are made of 100 words for Mboshi and 60 words for Kunwinjku. We randomly selected words which occur at least 3 times. For each word, we manually selected examples clearly pronounced, verified and respecting the speaker distribution of the corpora (Table~\ref{tab:spk_kun} and \ref{tab:spk_mb}) and clipped them out of the corpora used for spoken term detection. 

\begin{table}[h!]
\begin{tabular}{l|l|l|l|l|l}

Speaker      & RB & TG & GN & SG & MM \\ \hline
Distribution & 10\%   & 25\%   & 15\%  & 38\%   & 12\%  
\end{tabular}
\caption{Speaker distribution across Kunwinjku lexicon}
\label{tab:spk_kun}
\end{table}

\begin{table}[h!]
%\begin{center}
\begin{tabular}{l|l|l|l}

Speaker      & Abiayi & Kouarata & Martial \\ \hline
Distribution & 63\%   & 33\%     & 4\%     \\ 
%\end{center}
\end{tabular}
\caption{Speaker distribution across Mboshi lexicon}
\label{tab:spk_mb}
\end{table}

\section{Results}
\label{sec:results}

We first evaluate the phone error rate (PER) for both languages.
%with the gold standard in Mboshi and the gold standard mapped to phones in Kunwinjku. 
For Kunwinjku the PER started at 53.15\%, and we obtained 40.55\% after the system early stopped at the 17th epoch.
For Mboshi the PER started at 59\% and reached 35.49\% at the 29th epoch.
Although the PER is low considering the small amount of data used for fine-tuning Allosaurus, we would expect a bigger difference between Kunwinjku and Mboshi considering that Mboshi is read speech without foreign words and Kunwinjku is spontaneous speech containing English words.

We evaluate the two spoken term detection approaches presented in previous section, using recall and precision. %In other words, we want to know between the words candidates output by out two methods, how many are actually present in the orthographic gold transcription and how many are still to be retrieved. 
%The recall is defined as the number of items retrieved from the full corpus $X$ out of the number of all retrievable items in the corpus, i.e., the intersection of the lexicon $L$ and the corpus $C$, $X/(L \cap C)$.
%In other words, this recall corresponds to the coverage of the lexicon related to its tokens in the speech collection.
%The recall is defined as the number of true positive retrieved on the intersection of the lexicon and the corpus and 
%The precision is the number of true positive on the total number of items retrieved. 
Since the recall of the baseline includes the words from the lexicon used as queries, we also provide the recall without the instances of the lexicon (recall-no-lex).
The F-score is computed using regular recall.
In the case of Kunwinjku, the size of the validation set is small, which means that the words from the lexicon represent a large portion of the final results.
% \begin{table}[!h]
% \begin{tabular}{c|c|c|c|c}
%          & recall-no-lex & recall  & precision & F-score  \\\hline

% baseline & 32.71\% & 48.67\% & 6.83\% & 11.98\%\\ \hline
% P2W & - & 18.08\% & 80\% & 29.49\%
% \end{tabular}

% \caption{score word spotting Kunwunjku}
% \label{tab:score_kun}
% \end{table}

% \begin{table}[!h]
% \begin{tabular}{c|c|c|c|c}
%          & recall-no-lex &recall  & precision & F-score   \\\hline

% baseline & 21.97\% & 27.68\% & 7.7\% & 12.05\% \\ \hline
% P2W & - & 16.49\% & 62.28\% & 26.07\%
% \end{tabular}
% \caption{score word spotting Mboshi}
% \label{tab:score_mb}
% \end{table}
\begin{table}[!h]
\begin{tabular}{c|c|c|c|c}
         & recall-no-lex & recall  & precision & F-score  \\\hline

DTW & 17.43\% & 33.24\% & 21.67\% & 26.24\%\\ \hline
P2W & - & 15.15\% & 85.07\% & 25.72\%
\end{tabular}

\caption{Performance of spoken term detection in Kunwunjku}
\label{tab:score_kun}
\end{table}

\begin{table}[!h]
\begin{tabular}{c|c|c|c|c}
         & recall-no-lex &recall  & precision & F-score   \\\hline

DTW & 15.47\% & 21.18\% & 13.55\% & 16.53\% \\ \hline
P2W & - & 13.07\% & 62.91\% & 21.64\%
\end{tabular}
\caption{Performance of spoken term detection in Mboshi}
\label{tab:score_mb}
\end{table}

Results are detailed in Table~\ref{tab:score_kun} for Kunwinjku and in Table~\ref{tab:score_mb} for Mboshi.
DTW has an advantage over P2W that it allows us to look deeply into the speech collection and so to retrieve more items and increase the recall, while decreasing precision.

P2W allows us to navigate into a limited number of symbols and matched words from an initial lexicon. %into the automatic graphs transcription. 
This method allows us to navigate only in the stream of symbols automatically generated which  makes the word matching results highly dependant of the phones recognition performance. However, navigating into written forms allows very accurate matches and a high precision since we are matching only exact same forms between the lexicon and the unsegmented stream of characters. The lower precision for Mboshi can be explained by the fact that this language has tones which increases the number of confusable vowels.

\cite{le2020enabling} pointed out the limitation of their method in terms of cross speaker spoken term detection. 
%To demonstrate the better robustness (to speaker change) of the phone recognition approach, we made the following  analysis. 
To compare the two approaches on this aspect, we analyze each true positive output:
%For each true positive output, 
we check if the word matched is pronounced by a same or different speaker that the query term. Even if we only use the written forms of the queries for P2W, we also make the same analysis.
\begin{figure}[!ht]%
    \includegraphics[width=0.45\textwidth]{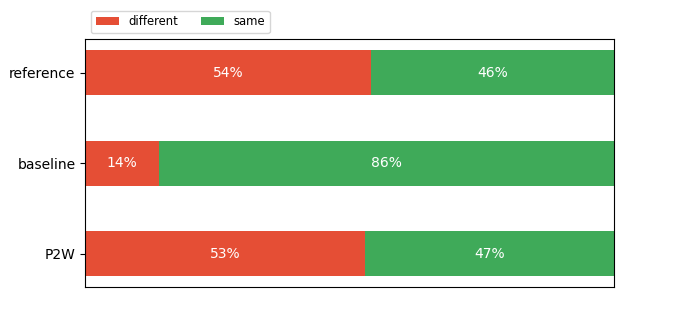}%
    \caption{ Proportion of same-speaker/different-speaker retrieval in Kunwinjku}%
    \label{fig:speaker_kun}%
\end{figure}
\begin{figure}[!ht]%
    \includegraphics[width=0.45\textwidth]{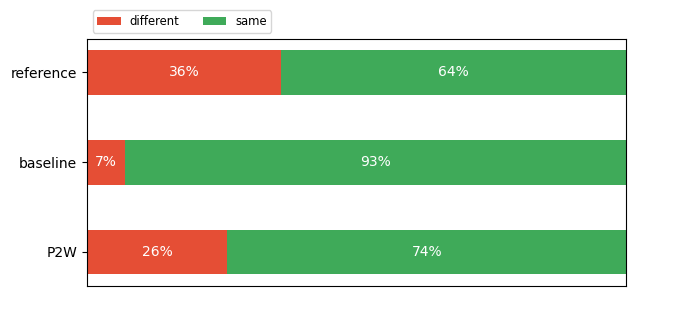}%
    \caption{ Proportion of same-speaker/different-speaker retrieval in Mboshi  }%
    \label{fig:speaker_mb}%
\end{figure}

Figures \ref{fig:speaker_kun} and \ref{fig:speaker_mb} present the proportion of retrieved spoken term from same-speaker or different-speaker. For a fair comparison, we also compute the distribution of same/different speaker between the lexicons and the words to be retrieved in the corpora (reference). We can see from Figures \ref{fig:speaker_kun} and  \ref{fig:speaker_mb} that P2W method follows the general distribution in the corpora while the baseline DTW retrieves mostly terms pronounced by the same speaker.

\begin{figure}[!ht]%
    \begin{center}
    \includegraphics[width=0.4\textwidth]{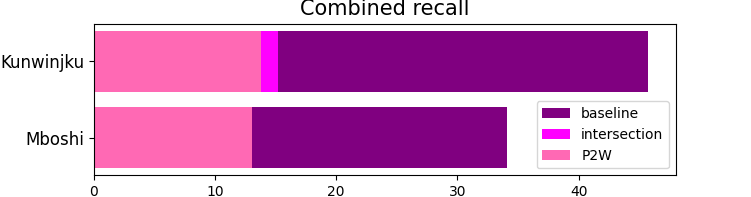}%
    \caption{Relative coverage of the combined methods, the recall of the baselines includes the spoken lexicon}%
    \label{fig:speaker}%
        \end{center}
\end{figure}

We mentioned in Section \ref{sec:context} that DTW and phone recognition have their own strength. DTW will cope easily with spontaneous speech and sounds elision due to co-articulation effects. Phone recognition allows us to avoid gathering spoken query and retrieve terms with exact matching between written forms. 
To highlight the complementary  of the two methods,  we analyse the intersection of their true positives in figure \ref{fig:speaker}. We show that for both corpora the intersection of the true positives are small and combining the two methods can help us increase the coverage of the transcription. 

\section{Taking advantage of phoneme transcription ambiguity}
\begin{algorithm}
\begin{algorithmic}
\State $token \gets None$
\State $valid \gets None$
\State $i \gets 0$
\While{$i \leq confnet[0].length$}
    \ForEach{$phone \in confnet[i]$}
        \If {$confnet[i][phone] \in lexState.current.validPath$}
            \State $token \gets token + confnet[i][phone]$
            \State $lexState.current \gets confnet[i][phone]$
            \If {$lexState.current.isword$} $valid \gets token$ \EndIf
            \If {$savePoint = None$} $savePoint \gets i$ \EndIf
            \State \textbf{break}
        \Else
            \If{$valid \neq token$} 
            \State $index \gets savePoint$ 
            \State $savePoint \gets None$
            \EndIf
            \State $lexState.current \gets lexState.root$
            \State $token \gets None$
            \If{$valid \neq None$} \Return $valid$ \EndIf
        \EndIf
    \EndFor
    \State $i \gets i + 1$
\EndWhile
\end{algorithmic}
\label{algo}
\caption{search of the phoneme confusion network}
\end{algorithm}

\begin{figure}[!ht]%
    \centering
    \subfloat[lexicon]{\label{sub:trie}{\includegraphics[width=0.08\textwidth]{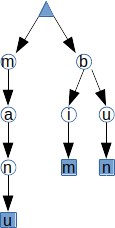}}}%
    \hspace{3mm}
    \subfloat[confusion network]{\label{sub:matrix}{\includegraphics[width=0.25\textwidth]{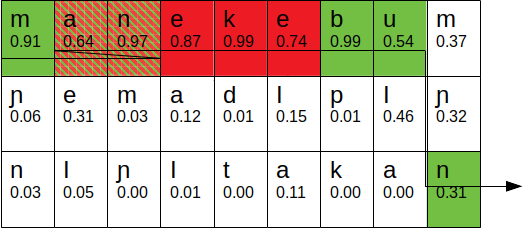}}}
    \caption{search of the phoneme confusion network \ref{sub:matrix} with lexicon stored in a trie \ref{sub:trie}}%
    \label{fig:topk}%
\end{figure}

Allosaurus allows us to output a confusion network \cite{DBLP:journals/csl/ManguBS00} of phonemes instead of a simple phoneme transcription sequence. This graph contains the top $k$ hypotheses for each phone in the sequence, associated with a probability (Fig.~\ref{sub:matrix}).
In order to push the search further, we incorporate the words of the lexicon in a trie and implement a simple search (see Algorithm~1). The idea is to check in the confusion network ($confnet$) for each top $k$ phone from the most likely to the least likely if it corresponds to a valid path in the lexicon ($lexState$).
If a given phone is recognized as a final state in the lexicon, the corresponding word is kept as a candidate.
If a followed path cannot find a final state, the search starts back from the index corresponding to the first state recognized.
This is a basic search which allows us to easily explore the network but could be optimized following a Viterbi algorithm adapted for this task. An illustration of the search is shown in Figure \ref{fig:topk} where we start searching for \textit{manu}. The search starts back from \textit{a} since no final state has been found. No phone is recognized as a valid state until \textit{b}. The search ends recognizing \textit{bun}.
%and parse the confusion network as follows (see Figure \ref{fig:topk}): we check for each character in the 
% most probable sequence if it corresponds to a valid transition in the trie. If a transition is not found the least probable sequences are checked, (from the most probable to the least probable ones).
% If a character corresponds to the last state of an element in the trie, the word is validated as candidate. If a followed path does not end up to a valid word, \textbf{the parsing get back to the index corresponding to the second letter of the word recognized in the network} \todo{[unclear what it means exactly]}. In the example illustrated in 
% \todo{[This part needs major rewriting (including what follows)]} Figure \ref{fig:topk}, we start searching for \textit{manu}, we get back to \textit{a} since the parsing did not reach a final state. We cannot find a correct path until \textit{b}. We then finish the parsing finding \textit{bun}.

We applied this search on the top 5 phonemes applying a pruning of the network based on two arbitrary thresholds (0.2 and 0.1).
We then compute the same evaluation described in Section~\ref{sec:results} and report the results in Tables~\ref{parsing_kun} and \ref{parsing_mb}. 
\begin{table}[!ht]
% \begin{tabular}{c|c|c|c}

%       & recall & precision & F-score \\ \hline
% baseline & 33.24\% & 21.67\% & 26.24\%\\ \hline
% P2W & 15.16\%     & 85.07\%  & 25.72\%  \\ \hline
% P2W 0.2   & 18.35\%     & 75.82\%  & 29.55\%  \\ \hline
% P2W 0.1   & 22.60\%     & 70.83\%  & 34.27\%  \\ 
% \end{tabular}

\begin{tabular}{c|c|c|c|c}

      & recall-no-lex & recall & precision & F-score \\ \hline
baseline & 17.43\% & 33.24\% & 21.67\% & 26.24\%\\ \hline
P2W & - & 15.16\%     & 85.07\%  & 25.72\%  \\ \hline
P2W 0.2 & -  & 19.15\%     & 74.23\%  & 30.45\%  \\ \hline
P2W 0.1 & - & 26.60\%     & 65.35\%  & 37.81\%  \\ 
\end{tabular}

\caption{Results of the search for Kunwinjku}
\label{parsing_kun}
\end{table}

\begin{table}[!ht]

\begin{tabular}{c|c|c|c|c}

      & recall-no-lex & recall & precision & F-score \\ \hline
baseline & 15.47\% & 21.18\% & 13.55\% & 16.53\%\\ \hline
P2W & - & 13.07\%     & 62.91\%  & 21.64\%  \\ \hline
P2W 0.2 & - & 18.77\%     & 58.23\%  & 28.39\%  \\ \hline
P2W 0.1 & -  & 23.74\%     & 47.54\%  & 31.67\%  \\ 
\end{tabular}
\caption{Results of the search for Mboshi}
\label{parsing_mb}
\end{table}
We can see that a simple search allows us to boost the recall maintaining the precision at an acceptable rate. We can also see that P2W with a pruning set at 0.2 outperforms the DTW baseline in terms of recall when the spoken lexicon is not included in the results.

\section{Conclusion}

Although traditional methods of spoken term detection are more robust in spontaneous speech and cope more easily with speech elision due to coarticulation effects, the phone recognition based approach appears to be much more accurate mostly because of its speaker independence.
Even if cleaner data is to be privileged for fine-tuning the model for better recall, a noisy transcription can still be usable.
In the context of interactive transcription, precision is to be privileged over recall in order to properly assist the transcriber limiting the correction of wrong output.

The spoken lexicon required by DTW can be hard to collect:
we saw that differences of speaker between the corpus and the lexicon have a impact on the final precision which force us to carefully select the example to be used and extract them manually.
A phone recognition based approach would only need a written lexicon.

Phones recognition based methods give us the possibility of quickly and precisely exploring the content of a given speech collection through simple textual queries.
The confusion network output by Allosaurus lets us enable a deeper search, replacing a given phone by another.
Such exploration outperforms the DTW baseline in terms of recall, while maintaining high precision.

\section{Acknowledgements}

We are grateful to the Bininj people of Northern Australia for the opportunity to work in their community,
and particularly to artists at Injalak Arts and Craft (Gunbalanya) and to the Warddeken Rangers (Kabulwarnamyo, Manmoyi). This research was covered by a research permit from the Northern Land Council, ethics approved from CDU and was supported by the Australian government through a PhD scholarship, and grants from the Australian Research Council and the Indigenous Language and Arts Program.
% \begin{table}[!h]
% \begin{tabular}{l|l|l|l}
%          & recall-no-lex & recall  & precision   \\\hline
% baseline & 33.59\%&42.84\% & 20.82\%  \\\hline
% kws\_1\_ite & 28.35\%& 37.59\% & 10.22\%  \\ \hline
% kws\_lex2& 35.37\% & 51.33 & 12.04\% \\ \hline
% kws\_lex2\_1\_ite & 32.71\% & 48.67 & 6.83\% \\ \hline
% P2W      & -&23.57\% & 81.81\% \\ \hline
% P2W\_lex2 & - & 18.88\% & 78.02\%
% \end{tabular}
% \caption{score word spotting Kunwunjku}
% \end{table}

%   \begin{table}[!h]
% \begin{tabular}{l|l|l|l}
%          & recall-no-lex &recall  & precision   \\\hline
% baseline & 21.57\% &24.42\% & 30.34\%   \\\hline
% kws\_1\_ite & 19.56\% & 22.40\% & 13.78\% \\ \hline
% kws\_lex2& 18.83\% & 24.54 & 17.80\% \\ \hline
% kws\_lex2\_1\_ite & 21.97\% & 27.68\% & 7.7\% \\ \hline
% P2W      & - &18\% & 76.11\%  \\ \hline
% P2W\_lex2 & - & 16.49\% & 62.28\%
% \end{tabular}
% \caption{score word spotting Mboshi}
% \end{table}
% Mboshi at 35.49\% PER. Kunwunjku at 40.55\% PER
% \begin{itemize}
%     \item compare the PER before and after training
%     \item eval of different methods of segmentation (\textcolor{red}{I don't have the words bounderies on the automatic transcription, but I was thinking I can generate it with some kind of alignment with the gold standard?}
%     \item compare recall/precision between the new pipeline and coling with the final 60words lexicon
%     \item run the segmentation with the lexicon extracted from the bible and njamed and eval with WER.
% \end{itemize}

\vfil\pagebreak

\bibliographystyle{IEEEtran}

\bibliography{mybib}

% \begin{thebibliography}{9}
% \bibitem[1]{Davis80-COP}
%   S.\ B.\ Davis and P.\ Mermelstein,
%   ``Comparison of parametric representation for monosyllabic word recognition in continuously spoken sentences,''
%   \textit{IEEE Transactions on Acoustics, Speech and Signal Processing}, vol.~28, no.~4, pp.~357--366, 1980.
% \bibitem[2]{Rabiner89-ATO}
%   L.\ R.\ Rabiner,
%   ``A tutorial on hidden Markov models and selected applications in speech recognition,''
%   \textit{Proceedings of the IEEE}, vol.~77, no.~2, pp.~257-286, 1989.
% \bibitem[3]{Hastie09-TEO}
%   T.\ Hastie, R.\ Tibshirani, and J.\ Friedman,
%   \textit{The Elements of Statistical Learning -- Data Mining, Inference, and Prediction}.
%   New York: Springer, 2009.
% \bibitem[4]{YourName17-XXX}
%   F.\ Lastname1, F.\ Lastname2, and F.\ Lastname3,
%   ``Title of your INTERSPEECH 2020 publication,''
%   in \textit{Interspeech 2020 -- 20\textsuperscript{th} Annual Conference of the International Speech Communication Association, September 15-19, Graz, Austria, Proceedings, Proceedings}, 2020, pp.~100--104.
% \end{thebibliography}

\end{document}